\documentclass[letterpaper, 10 pt, conference]{ieeeconf}  

\usepackage{textcomp}
\usepackage{amsfonts,amssymb} 
\usepackage{booktabs}
\usepackage{stfloats}
\usepackage{url}
\usepackage{verbatim}
\usepackage{graphicx}
\usepackage{cite}
\usepackage[version=4]{mhchem}
\usepackage{graphicx} 
\usepackage{epsfig} 
\usepackage{mathptmx} 
\usepackage{times} 
\usepackage{amsmath} 
\usepackage{amssymb}  
\usepackage[T1]{fontenc}
\usepackage{tabularx}  
\usepackage{hyperref}
\hypersetup{hidelinks,
	colorlinks=true,
	allcolors=black,
	pdfstartview=Fit,
	breaklinks=true}
\usepackage{multirow}
\usepackage{url}
\usepackage{makecell}
\usepackage{diagbox} 
\usepackage{bm}
\usepackage[linesnumbered,ruled,vlined]{algorithm2e}

\IEEEoverridecommandlockouts                              
\title{\LARGE \bf
Dual-modal Tactile E-skin: Enabling Bidirectional Human-Robot Interaction via Integrated Tactile Perception and Feedback
}

\author{Shilong Mu\textsuperscript{*}, Runze Zhao\textsuperscript{*}, Zenan Lin, Yan Huang, Shoujie Li, Chenchang Li,\\ Xiao-Ping Zhang,~\IEEEmembership{Fellow,~IEEE}, Wenbo Ding
\thanks{*These authors contributed equally to this work.}
\thanks{This work was supported by Shenzhen Ubiquitous Data Enabling Key Lab under Grant No. ZDSYS20220527171406015, by Guangdong Innovative and Entrepreneurial Research Team Program (2021ZT09L197), by Shenzhen Science and Technology Program (JCYJ20220530143013030), by Tsinghua Shenzhen International Graduate School-Shenzhen Pengrui Young Faculty Program of Shenzhen Pengrui Foundation (No. SZPR2023005). We also acknowledge the support from the Tsinghua Shenzhen International Graduate School-Shenzhen Pengrui Endowed Professorship Scheme of Shenzhen Pengrui Foundation. (Corresponding author: Wenbo Ding, ding.wenbo@sz.tsinghua.edu.cn) }
\thanks{Shilong Mu, Runze Zhao, Zenan Lin, Shoujie Li, Chenchang Li, Xiao-Ping Zhang, Wenbo Ding are with Tsinghua-Berkeley Shenzhen Institute, Shenzhen International Graduate School, Tsinghua University, Shenzhen, China, 518055.}
\thanks{Yan Huang is with the School of Electronic Information, Wuhan University, Wuhan, China.}%
\thanks{Xiao-Ping Zhang is also with the Department of Electrical, Computer and Biomedical Engineering, Ryerson University, Toronto, ON M5B 2K3, Canada.}
\thanks{This paper has supplementary downloadable material available at: https://sites.google.com/view/touch-e-skin/.}
}

\begin{document}
\maketitle
\thispagestyle{empty}
\pagestyle{empty}

\begin{abstract}
To foster an immersive and natural human-robot interaction (HRI), the implementation of tactile perception and feedback becomes imperative, effectively bridging the conventional sensory gap. In this paper, we propose a dual-modal electronic skin (e-skin) that integrates magnetic tactile sensing and vibration feedback for enhanced HRI. The dual-modal tactile e-skin offers multi-functional tactile sensing and programmable haptic feedback, underpinned by a layered structure comprised of flexible magnetic films, soft silicone elastomer, a Hall sensor and actuator array, and a microcontroller unit. The e-skin captures the magnetic field changes caused by subtle deformations through Hall sensors, employing deep learning for accurate tactile perception. Simultaneously, the actuator array generates mechanical vibrations to facilitate haptic feedback, delivering diverse mechanical stimuli. Notably, the dual-modal e-skin is capable of transmitting tactile information bidirectionally, enabling object recognition and fine-weighing operations. This bidirectional tactile interaction framework will enhance the immersion and efficiency of interactions between humans and robots.
\end{abstract}

\section{Introduction}

\begin{figure}[htbp]
    \centering
    \includegraphics[width=0.45\textwidth]{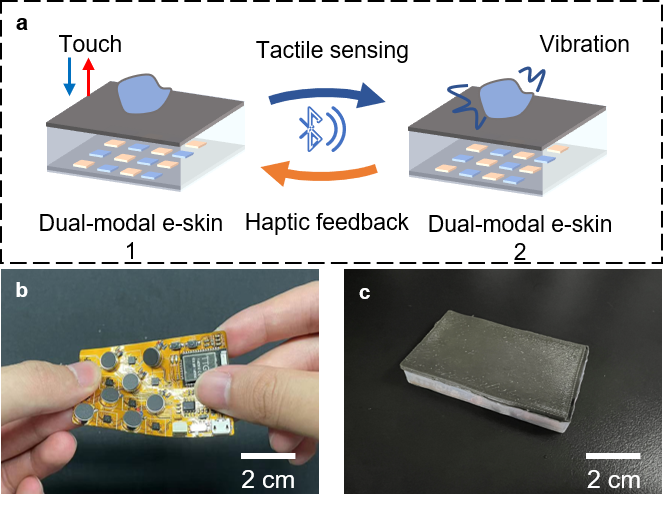}
    \caption{(a) Schematic diagram of dual-modal e-skin bidirectional tactile interaction. (b) Picture of the fexible printed circuit board (FPCB) with functions of tactile sensing and actuation and Bluetooth module. (c) Photo of dual-modal e-skin.}
    \label{fig1}
\end{figure}

Inspired by the multimodal sensory capabilities of biological skin, electronic skin (e-skin) has been meticulously designed and developed to be affixed to the human epidermis~\cite{lopes2018hydroprinted,niu2022perception}, prosthetic limbs~\cite{zhang2022wide,wu2018skin}, or robotics~\cite{mu2023platypus,kim2022dynamic,9376677} to provide a range of sensing and feedback functions. Owing to the excellent stretchability~\cite{zhang2022stretchable} and high biocompatibility~\cite{gong2020biocompatible}, e-skin has clearly demonstrated its advantages across multiple fields, including biomedical engineering~\cite{zhu2020flexible}, healthcare monitoring~\cite{hu2023stretchable}, and human-robot interaction (HRI)~\cite{liu2022electronic,9246727}. Furthermore, e-skin not only imitates the tactile sensory capabilities of human skin but also provides diverse haptic feedback modalities, thereby enriching the immersive experience. Dual-modal e-skin, endowed with both tactile sensing and haptic feedback capabilities, has substantially promoted more natural and immersive interactions between humans and robots, or humans themselves, laying the foundation for more challenging collaborative tasks~\cite{li2022touch,song2021miniaturized,park2023skin}.

Unfortunately, current e-skin technologies often only have single-function capabilities for either tactile perception or haptic feedback. Through e-skin based on piezoresistive~\cite{zhang2023localizing,ruoqin2023miura}, capacitive~\cite{huang2022high,zhang2023flexible,10160961}, optical waveguide~\cite{ren2023mechanoluminescent,zhao2023review}, triboelectric~\cite{wang2023stev} and other mechanisms~\cite{2022Guiding,2022A}, external stimuli can be converted into electrical signals or other forms of data, thereby laying the foundation for abundant tactile sensory perception. At the same time, e-skin can also provide haptic feedback through mechanisms such as mechanical vibration~\cite{sonar2021soft,9341087}, electrical stimulation~\cite{2022Encoding,doi:10.1126/sciadv.abp8738}, and temperature changes~\cite{doi:10.1126/scirobotics.abl4543,10049651}. These feedback mechanisms can simulate tactile properties such as force, vibration, and texture, allowing users to perceive virtual touch in a more realistic way. However, because the working mechanisms of the perception and feedback units cannot be seamlessly combined, resulting in larger devices and higher manufacturing costs, there are still huge challenges in the development of e-skins that simultaneously achieve tactile sensing and feedback.

\begin{figure*}[thpb]
  \centering
  \includegraphics[width=1.0\linewidth]{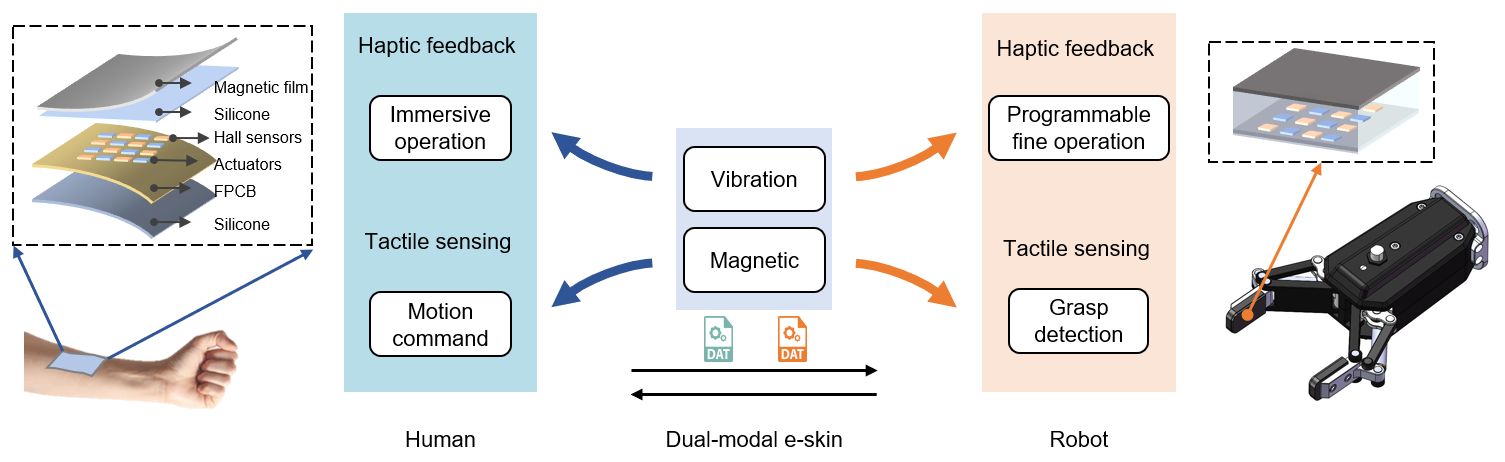}\\
  \caption{Framework diagram of bidirectional tactile human-robot interaction. The dual-mode e-skin is placed on the human arm and the gripper of the robot respectively, and transmits tactile sensing and vibration feedback information bidirectionally and wirelessly.}
  \label{fig2}
\end{figure*}

In this study, we present an integrated e-skin that combines magnetic tactile sensing with vibration feedback (as shown in Fig.~\ref{fig1}). Our goal is to provide a dual-modal tactile e-skin that offers multi-functional tactile perception, programmable haptic feedback, high scalability, and bidirectional wireless transmission of tactile information. These features are achieved through a layered structure composed of flexible magnetic films, silicone elastomer, an array of Hall sensors and vibration motors, and a microcontroller unit. The e-skin utilizes Hall sensors to detect the minute deformations of the flexible magnetic films caused by mechanical press. Haptic feedback is achieved through mechanical vibrations generated by actuator array. By integrating our dual-modal e-skin onto robotic or human skin, robots can acquire rich tactile sensing information in complex tasks, while humans can transmit commands through natural touch. The dual-mode e-skin can provide a variety of mechanical stimuli to the skin through programmable vibrations in the space and time domains, while enabling the robotic gripper to complete fine operations, such as fine and controllable metering of particles or powders. This bidirectional transmission of tactile information enhances the robot's perception and response capabilities, improving precise operation and efficient collaboration.

The main contributions of this work are as follows:
\begin{itemize}
\item Proposing an integrated e-skin that achieves multi-functional tactile sensing and programmable haptic feedback, addressing the limitation of previous single-modal e-skin.
\item Designing experiments to demonstrate the scalability of e-skin for applications in multiple fields, including grasping recognition and programmable fine operation.
\item Presenting a framework for bidirectional transmission of tactile information between humans and robots, demonstrating the potential for human-robot collaboration in complex tasks.
\end{itemize}

\section{Methods and Design}
\subsection{System framework overview}
The proposed bidirectional tactile interaction system between humans and robots is constituted by the deployment of dual-modal e-skin on both the human and the inner sides of the robotic grippers (as illustrated in Fig.~\ref{fig2}). Composed of a layered structure featuring magnetic films, silicone elastomer layers, and a flexible printed circuit board (FPCB) containing Hall sensor and an actuator array (4x4), along with a wireless transmission unit, the dual-modal e-skin exhibits a compact form factor with dimensions of 40x65 mm. The e-skin system is made of flexible materials and can be easily installed and integrated on human and robotic platforms. The e-skins at both ends transmit tactile information during the interaction process via wireless Bluetooth technology, covering data from eight three-axis Hall sensors for magnetic fields as well as vibration information from eight-channel actuators. On the human end, tactile interactions with the e-skin can wirelessly send motion control commands to the robot. Concurrently, the array of actuators employs a variety of mechanical vibrations to impose diverse stimulations on the human skin. On the robotic end, grippers equipped with the e-skin, utilizing magnetic tactile sensing amalgamated with deep learning algorithms, can recognize objects in real-time during grasping activities. Then the spatiotemporal programmable vibrations generated by the actuators can be transmitted to the terminal ends of the grasping tools, facilitating fine control mechanisms. The comprehensive human-machine bidirectional tactile interaction framework, which encompasses tactile sensing and programmable feedback modalities, is poised to significantly enhance the efficiency and immersion level of human-robot collaboration.

\begin{figure}[htbp]
    \centering
    \includegraphics[width=0.45\textwidth]{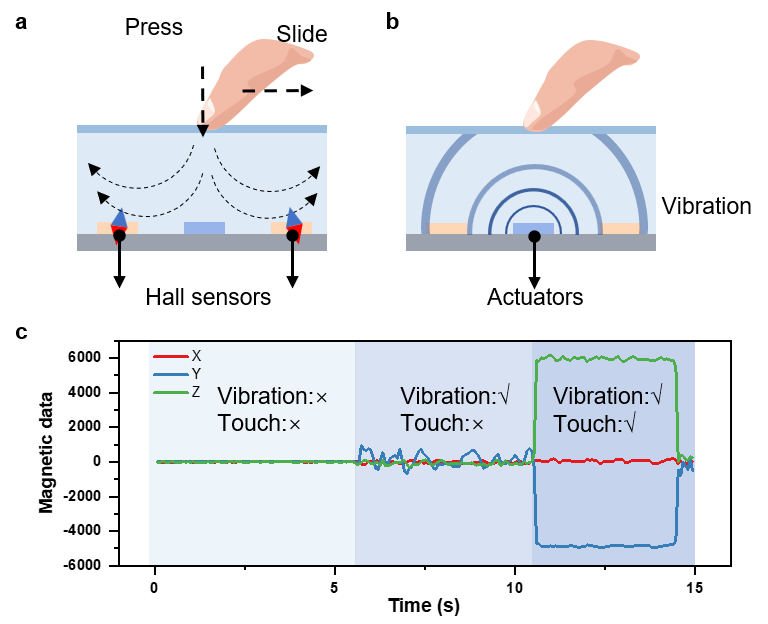}
    \caption{(a-b) Working principle of the proposed dual-modal e-skin. (c) The impact of motor vibration on a single three-axis Hall sensor.}
    \label{fig3}
\end{figure}

\subsection{Working principle}

The dual-modal e-skin system employs a staggered array configuration, incorporating Hall sensors (QMC5883L) and compact vibrators with dimensions of 8 mm in diameter and 3 mm in height. This arrangement minimizes the overall system thickness to just 7 mm. The magnetized film within the system is highly responsive to external mechanical pressures and sliding motions. These external stimuli lead to alterations in the magnetic field, which are detected with high sensitivity by the three-axis Hall sensors. For ease of system expansion and integration, these Hall sensors communicate with a microcontroller (ESP32-PICO-D4) via an Inter-Integrated Circuit (I$^2$C) multiplexed connection. To ensure the reliability and accuracy of the collected data, the multi-channel sensor data are calibrated and zeroed before being wirelessly transmitted through Bluetooth technology.

In the actuator section, the system utilizes low-cost vibration motors that are notable for their small size, low power consumption, and cost-effectiveness. Through the implementation of multi-channel Pulse Width Modulation (PWM) control, these motors can operate efficiently within a supply voltage range of 0-3.7 V. Importantly, the actuators are affixed to flexible substrates and encapsulated in flexible materials, enabling their vibrations to be effectively conveyed to any external objects they come into contact with. 

In practical applications, the coexistence of magnetic tactile sensing and vibrational units within a unified system invariably amplifies design intricacies. This is particularly salient when considering that the magnetic fields emanating from the operational vibrational motors may engender perturbations that compromise the accuracy of the Hall sensors. To mitigate this challenge, we undertook rigorous experimental assessments and investigative studies. We first selected a Hall sensor as the test object, which was located near a magnetized film with a magnetization intensity of 2 mT. The experiment was divided into three stages to quantify the impact of the magnetic field generated by motor vibration on the Hall sensor. After careful testing and data analysis, we found that the influence exerted by the magnetic fields originating from motor vibrations on the Hall sensor was notably less significant than that imposed by a 4 N vertical pressure (Fig.~\ref{fig3}(c)). By optimizing the magnetization intensity, we can more effectively reduce the interference of the electromagnetic vibration motor, thus improving the stability of the entire system.

\begin{figure}[htbp]
    \centering
    \includegraphics[width=0.45\textwidth]{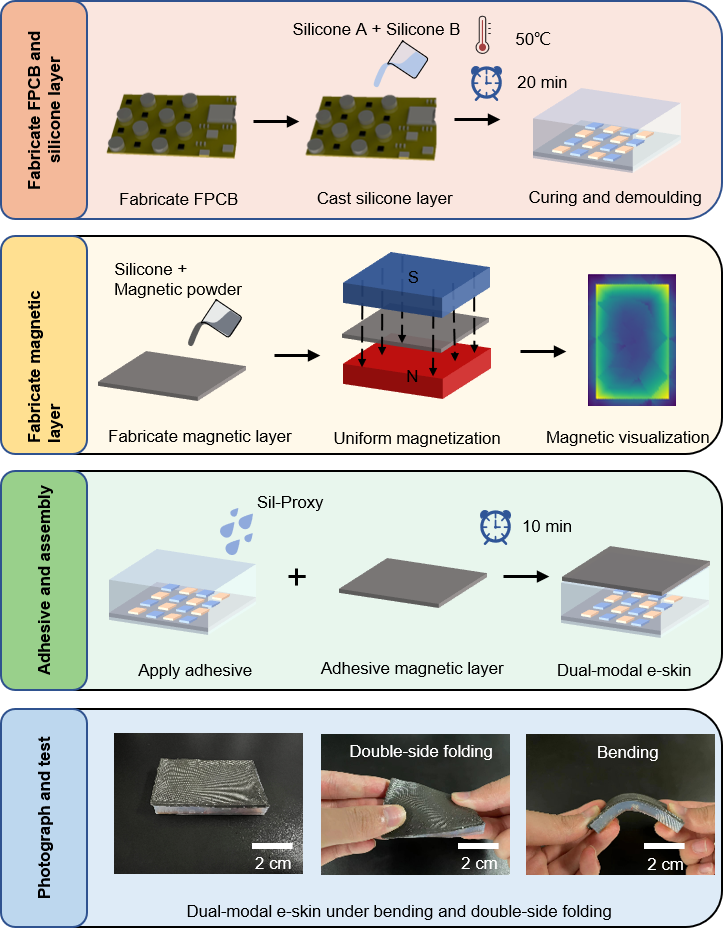}
    \caption{Fabrication process of the dual-modal e-skin.}
    \label{fig4}
\end{figure}

\subsection{Dual-modal e-skin fabrication}

As shown in Fig.~\ref{fig4}, we successfully fabricated a dual-mode e-skin with a flexible multi-layer structure. The magnetized film has a thickness of 1.5 mm, and it is paired with a 4.5 mm-thick silicone layer (Shenzhen Hongye Silicone, E600) to completely cover the FPCB. Furthermore, we tailored the material's elastic modulus through judicious adjustment of the silicone-to-additive ratio, thereby customizing its mechanical properties to meet the exigencies of individual application scenarios.

Then we mixed liquid silicone and neodymium (NdFeB) magnetic powders with a weight ratio of 1:1 and stirred thoroughly to ensure that the magnetic powder was evenly distributed in the silicone mixture. The mixture was then injected into a pre-made mold printed from polylactic acid (PLA) material and cured. In the curing process, we used vacuum equipment to remove air bubbles from the mixture and then allow it to solidify naturally.

The magnetization of the magnetized film was completed by a special magnetizing machine (PS-DTC60, Hunan Paisheng Technology). After magnetization, we used a high-precision magnetic field meter to perform array scanning and visually analyze the magnetic field distribution of the magnetized film. Then the FPCB with silicone layer was coated with silicone adhesive (Sil-Proxy, Smooth-On) and adhered to the magnetized film. After about 10 minutes of curing time, the dual-mode e-skin is completed.

Finally, we conducted a series of flexibility tests on the e-skin, including bending down and double-sided folding tests, and the results showed that it has excellent flexibility and adaptability, and can be applied to various parts of the human body and robots.

\begin{figure*}[thpb]
  \centering
  \includegraphics[width=1.0\linewidth]{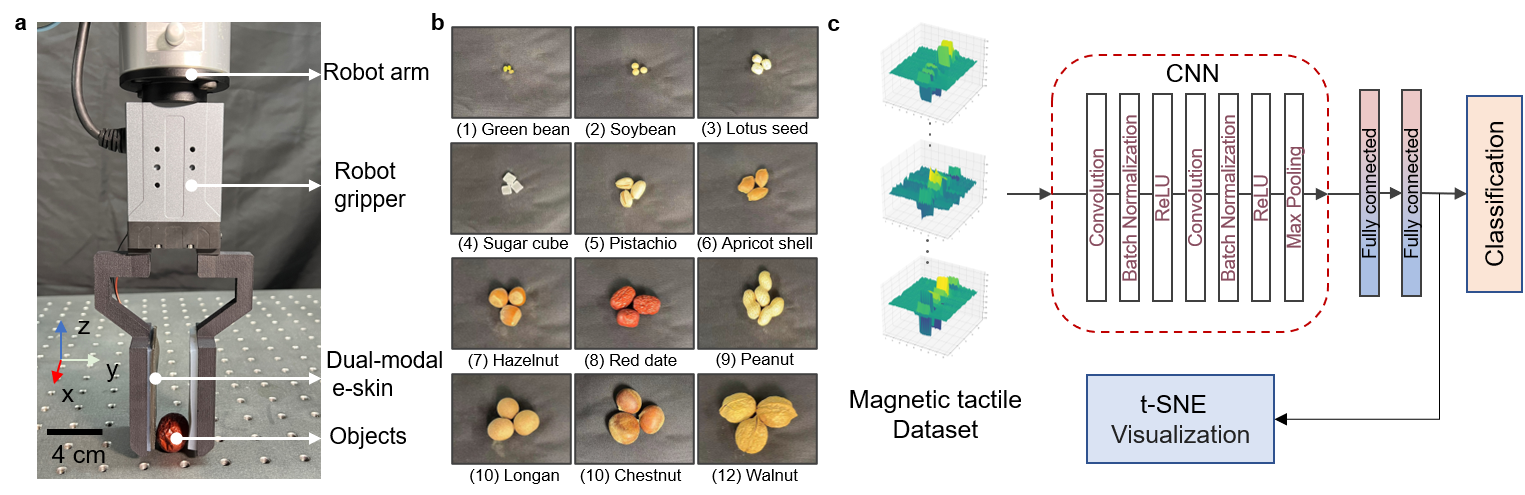}
  \caption{Dual-modal e-skin-enabled grasping recognition. (a) Schematic diagram of a robot gripper assembling an e-skin. (b) Twelve types of common objects. (c) Array tactile information processing flow chart.
}\label{fig5}
\end{figure*}

\subsection{Device cost}

Existing sensing and actuating array systems often require complex wiring and expensive material preparation. Furthermore, the integration of too many rigid components further inflates the cost and complicates the deployment process. Compared to existing dual-modal e-skins, our proposed device is more cost-effective while still maintaining ease of fabrication and material flexibility.

Leveraging off-the-shelf electronic components reduces manufacturing costs while enhancing system stability. Additionally, we utilized a three-dimension (3D) printer to create the curing mold, at a cost of approximately one US dollar. As can be seen in Table 1, the total cost of the device is less than 26 US dollars, and it weighs less than 29 grams. This highlights the system's affordability and lightweight nature, offering a significant advantage over conventional solutions.

\begin{table}[h]
 \centering
 \caption{Cost and Weight Details of the Proposed Device}
 \label{tab:pagenum}
 \begin{tabular}{l | c | c | c}
  \hline \hline
  & Number & Cost (USD)  & Weight (g) \\
  \hline
FPCB  & 1   & 20  &    4   \\
  Vibration motor  & 8 & 2.5   &  5   \\
  Silicone   & - & 1    &  14  \\
  Magnetic powder   & - & 1    &    4   \\
  Adhesive           & - & 0.5     &     2  \\
  3D printed parts           & - & 1     &     -  \\
  \hline
  Total                    & -   & 26   &  29 \\
  \hline \hline
 \end{tabular}
\end{table}

\section{Experiment}

Robots frequently encounter issues of obstructed vision and low visibility when engaged in human-robot collaborative tasks. Relying solely on visual information can make object recognition and operation challenging. Additionally, low-frequency mechanical motion at the end of the robot may not be sufficient to perform some fine operations. To evaluate the capabilities of the dual-modal e-skin in tactile sensing and programmable feedback, we systematically designed three experiments: (1) A gripper equipped with the e-skin was used to classify objects based on tactile information.
(2) We controlled the weighing process of particles at the end of a spoon via a programmable vibrating array. (3) A bidirectional tactile interaction was employed to facilitate human-robot collaborative tasks.

\subsection{Grasping recognition}

\begin{figure}[htbp]
    \centering
    \includegraphics[width=0.45\textwidth]{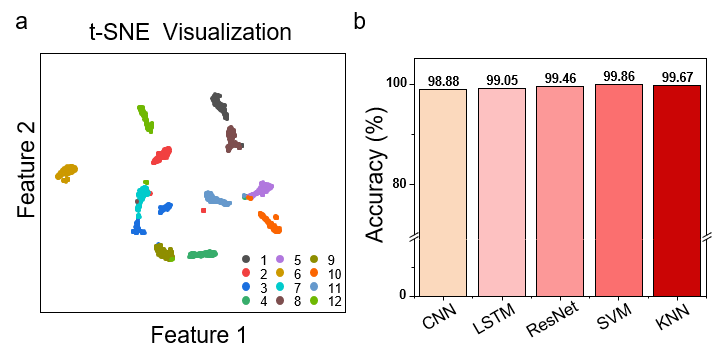}
    \caption{Recognition results and evaluation. (a) t-SNE clustering results. (b) Classification performance comparison.}
    \label{fig6}
\end{figure}

To demonstrate the dual-modal e-skin's ability to capture subtle touch differences, we designed an experiment where we used a robot gripper to classify various objects through simple grasping (see Fig.~\ref{fig5} ). We equipped the gripper (PGE-15-26, DH-Robotics) at the end of a UR3 robot arm to perform repeated grasping tasks, while a dual-modal e-skin was affixed to one side of the gripper for tactile sensing and feedback. We chose 12 representative objects for the experiment, including green beans, sugar cubes, hazelnuts, and chestnuts, etc.

We defined the initial state where the gripper has not yet made contact with any object as the zero state. The gripper then slowly grasps each object from various angles for 0.2s and releases it, pausing for 3s before initiating a new grasp. For each object, we segmented the magnetic data from 8 channels into 200 samples. Finally, we randomly divided all the data samples (2400, 24, 60) into 1680 training samples and 720 test samples to train a classification model. Representative data samples and the data processing pipeline are shown in Fig.~\ref{fig5}(c).

\begin{figure*}[thpb]
  \centering
  \includegraphics[width=1.0\linewidth]{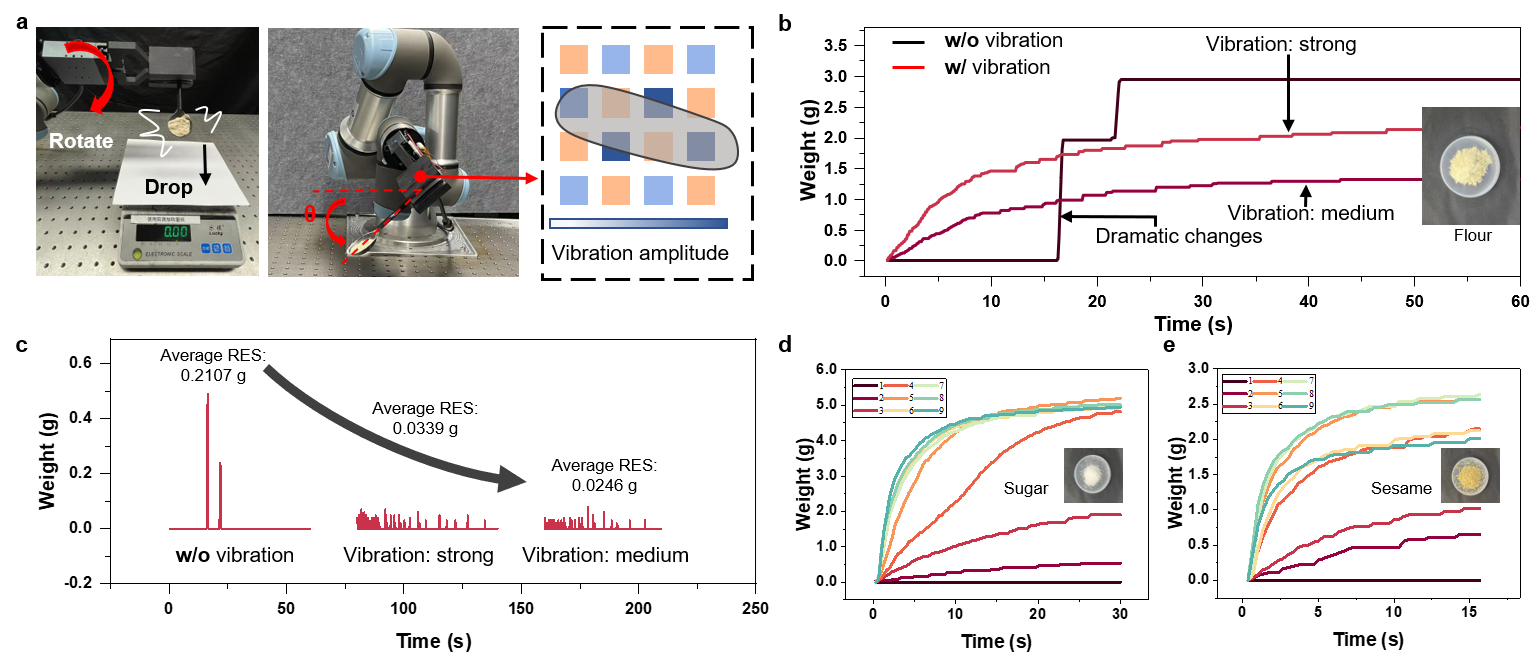}
  \caption{Programmable fine operation. (a) The programmable vibration array controls the uniform fall of powders and particles from the end of the clamping spoon. (b) Data comparison of weighing flour. (c) Difference curves and average resolution (RES) of weighed flour under different conditions (a = 1). (d-e) Control the dumping of different particles by adjusting the tilt angle and vibration intensity (d: white sugar, e: sesame). These labels (1-9) are identifiers for different curves on a graph, 1: tilting 30°and 2 motors vibrate, 2: tilting 30° and 4 motors vibrate, and so on.
}\label{fig7}
\end{figure*}

To achieve tactile-based object classification, we utilized a Convolutional Neural Network (CNN) architecture to extract features from tactile data, which resembles image-type data. The CNN proved to be effective even with a smaller sample size, achieving a training accuracy of 98.8\% by the 5th epoch. We used the t-SNE method to reduce the dimensionality and visualize results, so as to have an intuitive understanding of the distribution of recognition results. To verify the efficacy and generalizability of our tactile data, we conducted performance evaluations of the CNN model with alternative methodologies, specifically the Residual Network (ResNet), Long Short-Term Memory (LSTM) networks, etc. This comparative analysis further substantiates the robustness and versatility of using tactile data for object classification across different methodologies.

\subsection{Programmable weighing operation}

In fields ranging from scientific research and industrial production to pharmaceuticals, the precise weighing of powders and particles is of critical importance~\cite{carvalho2023automated}. Existing methods generally rely on either 1) rotating the spoon's tip to facilitate natural dropping of the powders and particles, or 2) using other tools to scrape off the material. However, some issues such as the formation of lumps make the operation highly sensitive, leading to poor repeatability and precision control.

To simulate this scenario, we selected three different types of solids (flour, white sugar, and sesame seeds) as the subjects for weighing. We designed an experimental platform, as shown in Fig.~\ref{fig7}(a). The gripper equipped with dual-mode e-skin was used to grasp the spoon's handle during the experiment. The falling of solid particles was studied by altering the vibrational state of the actuator array and the horizontal angle of the spoon. An electronic scale located below measured the weight of the falling solids in real-time.

As shown in Fig.~\ref{fig7}(b), we compared the flour weighing process with varying levels of programmable vibration amplitude (strong and medium) as opposed to no vibration. Our results indicate that without the implementation of e-skin for vibration control, the weight of the flour experiences a rapid increase during a 0-90° rotation. This can be attributed to the flour's tendency to clump in high-humidity environments, thereby compromising weighing accuracy. When different vibration amplitudes were activated, we observed varying rates of weight change in the flour, signifying that programmable vibrations can, to an extent, regulate weighing accuracy.

To quantify the change in mass over a given time interval, we introduce a quantifying metric \( \epsilon \), defined as follows:

\[
\epsilon = \frac{1}{N} \sum_{i=1}^{N} |m_{i+a} - m_i| \text{ where } m_{i+a} - m_i \neq 0,
\]

\begin{itemize}
    \item \( \epsilon \): Average change in mass, unit consistent with that of mass \( m \).
    \item \( m_i \): Mass at time point \( i \), where \( i = 1, 2, \ldots, n \).
    \item \( a \): Chosen time interval, a positive integer with \( a < n \).
    \item \( N \): Number of non-zero differences \( m_{i+a} - m_i \).
\end{itemize}

Our findings indicate that the average weighing resolution increases from 0.2107 g to 0.0246 g when different programmable vibration amplitudes are used, amounting to an improvement of nearly 8.5 times (see Fig.~\ref{fig7}(c)).

Furthermore, we examined the weighing impacts on white sugar and sesame seeds under nine different combinations: three rotational angles (30, 45, 50 degrees) and three vibration amplitudes (initiated with 50\% PWM for 2, 4, and 8 vibration motors).  During weighing, intuitively, higher vibration amplitudes and greater rotation angles shifted the weight curve to the left and increased the speed, as shown in Fig.~\ref{fig7}(c-d).

\subsection{Duplex human-robot immersion operation}

\begin{figure*}[htpb]
\centering
\includegraphics[width=1.0\linewidth]{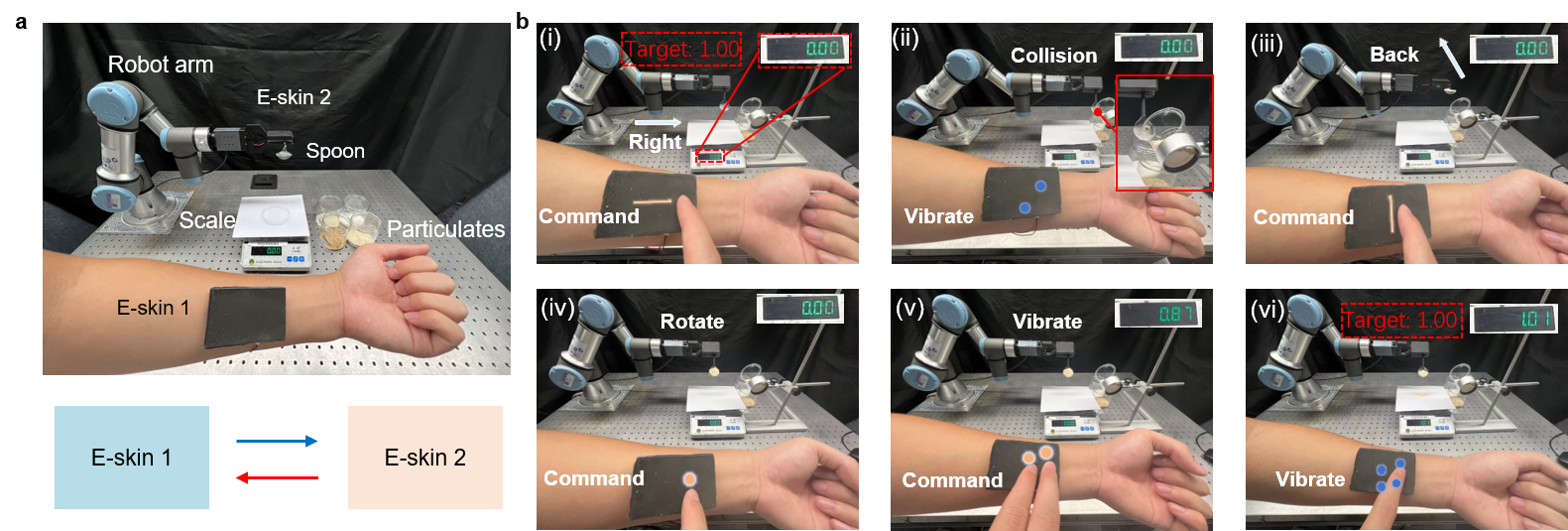}
\caption{Tactile perception and feedback experimental platform and bidirectional interaction process for immersive collaborative operations}
\vspace{-1.5em}
\label{fig8}
\end{figure*}

Human-machine interfaces (HMI) have shown great application potential in remotely operated robots, especially in areas such as immersive gaming, prosthetic or exoskeleton control, and high-precision operation. Although HMI systems and devices have advanced significantly, most are still limited by bulky mechanical structures and rigid electronic components, limiting the comfort and immersion of the user experience. To solve these problems, we designed a bidirectional tactile interaction system based on dual-modal e-skin.

In our experiment (shown in Fig.~\ref{fig8}(a)), two pieces of e-skin were attached to the human arm and the robot gripper respectively, and transmitted tactile data through wireless Bluetooth. Operators can remotely control a robotic arm equipped with a gripper and rely on real-time tactile commands and feedback to complete high-precision weighing operations. In Fig.~\ref{fig8}(b), by performing different forms of touch on the sensing area of the e-skin, such as pressing or sliding, the operator can convert complex motion commands into 8-channel tactile data, covering movement, grasping, and various vibration data. The microcontroller in the e-skin sends corresponding control commands to the PC controlling the robot through the serial port wired method. Next, the robot system executes the corresponding instructions. The entire bidirectional tactile interaction process can be divided into six stages, which include pre-setting a target weighing weight, followed by the operator applying different touch stimuli to the dual-modal e-skin to control the robot arm. Meanwhile, when a collision occurs at the end of the robot arm, the dual-modal e-skin detects it sensitively, and the e-skin on the human side generates vibrations, supplementing traditional visual feedback.

In conclusion, by controlling a programmable vibration array, the flour's weighing weight achieved was 1.01 g, closely approximating the target weight of 1.00 g. The experimental results indicate that the system not only enhances the robot's ability to operate in complex environments but also greatly improves the precision and reliability of operations through real-time tactile feedback.


\section{Discussion and Future Work}

In this study, we propose a wireless dual-mode e-skin capable of tactile sensing and haptic feedback. The e-skin can establish wireless bidirectional tactile interactions between humans and robots simultaneously, aiming to enrich HRI by providing an immersive and highly responsive interface. In order to verify the performance of dual-mode e-skin, we designed a grasping recognition experiment based on tactile information to verify the validity and universality of tactile data, and the object classification accuracy reached 98.88\%. In addition, the programmable vibration feedback array can control the speed of the fine-weighing process of particles and improve the control accuracy (\textasciitilde 0.0246 g). Finally, we established a human-robot bidirectional tactile interaction framework and system based on the proposed dual-mode e-skin. Experimental results show that the e-skin can effectively integrate multi-functional tactile sensing and feedback, and also opens up new ways for human-machine collaboration in complex tasks.

While our e-skin design has made promising strides, there are still some unexplored avenues for further research and enhancement. One immediate direction could be the miniaturization of e-skin components to allow easier integration into compact robotic systems or prosthetic devices. Interesting possibilities are to incorporate additional sensory modalities, such as temperature sensing and auditory feedback, to provide a more comprehensive sensory experience.

\bibliographystyle{ieeetr}
\bibliography{refe}
\end{document}